\documentclass{article}
\usepackage{spconf,amsmath,graphicx,multirow,float,url,cite,hyperref}

\title{Learning Deep and Compact Models for Gesture Recognition}
\name{Koustav Mullick and Anoop M. Namboodiri}
\address{Center for Visual Information Technology (CVIT), \\
International Institute of Information Technology,
Hyderabad, India.}

\begin{document}
%
\maketitle
\begin{abstract}
We look at the problem of developing a compact and accurate model for gesture recognition from videos
in a deep-learning framework. Towards this we propose a joint 3DCNN-LSTM model that is end-to-end
trainable and is shown to be better suited to capture the dynamic information in actions. The solution
achieves close to state-of-the-art accuracy on the ChaLearn dataset, with only half the model
size. We also explore ways to derive a much more compact representation in a knowledge distillation
framework followed by model compression.  The final model is less than $1~MB$ in size, which is less
than one hundredth of our initial model, with a drop of $7\%$  in accuracy, and is suitable for real-time
gesture recognition on mobile devices.
\end{abstract}
%
%
\section{Introduction and Related Work}
\label{sec:intro}

Gesture recognition is one of the key components in natural human-computer interfaces, especially for mobile devices, where interfaces like keyboard and mouse are impractical. However, the problem is challenging due to several factors like changes in background, lighting and variations in the performance and speed of the gestures. The users may have different appearance, pose and positioning relative to the camera as well as differences in the way they perform any specific gesture. In this work, we develop a solution that is accurate and robust enough to handle these variations, while being compact and fast to suit resource limited devices.

Deep Learning (DL) has had a major impact in the fields of computer vision, natural language processing and speech recognition over the past few years. However, one of the major issues with DL models are the large number of parameters involved which makes training time and memory requirements quite high. This makes it difficult to employ them in embedded devices. Keeping this in mind, we would like to explore ways to reduce the parameter space of such models without compromising a lot on performance.

\subsection{Action Recognition in Videos}
\label{ssec:action}

Human action in video sequences consists of spatial information within individual frames as well as temporal information of motion across the frames.
CNN models are very good at capturing the spatial information within an image as demonstrated by their success on ImageNet classification. Previous
works have tried to follow two approaches in integrating temporal information into these frameworks. The first one is to extract features from videos
that capture temporal information into images and then use image classification models. These include ideas such as dynamic image \cite{Bilen16a}
that produces a single image from a video obtained by applying rank pooling on the raw image pixels of a video, or the use of optical flow images~\cite{NIPS2014_5353}. 

A second approach is the use of deep-learning models that are better suited for capturing temporal information such as 3D-Convolutional Neural Network (3D-CNN)  \cite{Ji:2013:CNN:2412386.2412939} proposed by Ji \textit{et al.}, as well as Long Short Term Memory (LSTM) recurrent networks \cite{Hochreiter:1997:LSM:1246443.1246450}, which can use either pre-computed image features such as Bag of words and SIFT \cite{Baccouche:2010:ACS:1889001.1889024}, regular CNN features including optical-flow \cite{DBLP:journals/corr/NgHVVMT15} or 3D-CNN features \cite{Baccouche12spatio-temporalconvolutional}. Tran \textit{et al.} \cite{DBLP:journals/corr/TranBFTP14} proposed a simple, yet effective approach for spatio-temporal feature learning using deep 3D-CNN trained in a supervised fashion on a large scale video dataset. \cite{DBLP:journals/corr/NgHVVMT15} performed activity recognition by passing RGB and optical flow frames through pre-trained CNN networks and fed the learned features to an LSTM for final recognition. Varol \textit{et al.} \cite{varol16} learned video representation using 3D-CNN and demonstrated the impact of different low-level features as input.

Most of the works discussed above either use pre-trained CNN models for feature extraction or trains the CNN on parts of videos, separately from the model that performs actual classification (like LSTM or SVM). We would like to develop an end-to-end trainable model that can automatically learn both spatial and temporal features in a complete gesture. Towards this we combine the strengths of 3D-CNNs that are good at capturing spatial and short-term temporal features and LSTMs that can capture long-term temporal signatures of actions.

\subsection{Knowledge Distillation}
\label{ssec:distillation}

The second aspect that we explore in this work is to develop models that are efficient in terms of memory and speed.
Knowledge distillation, originally introduced by Caruana \textit{et al.} \cite{DBLP:conf/kdd/BucilaCN06}, introduced a process of transferring knowledge from one model to another. In order to obtain a faster inference, Hinton \textit{et al.} \cite{2015arXiv150302531H} proposed a compression framework which trains a student network, from the softened output of a larger model or an ensemble of wider networks, called teacher network. The idea is to allow the student network to capture not only the information provided by the true labels, but also the finer structure learned by the pre-trained teacher network. This approach has been used for speech recognition~\cite{DBLP:journals/corr/ChanKL15}, image classification~\cite{DBLP:journals/corr/RomeroBKCGB14}, and network compression~\cite{2015arXiv150302531H}.

In this work we show that the use of 3D-CNN with LSTM can improve the accuracy of action recognition in videos without increasing the model size, while the use of knowledge distillation allows us to develop a compact model that can be further compressed with minimal impact on accuracy.

\section{Our Approach}
\label{sec:approach}

We describe our framework in three stages. As baseline models we use a 3D-CNN and an LSTM variant of RNN to classify each gesture. For 3D-CNN each gesture video needs to be represented using a fixed number of frames. For LSTM the raw frames are flattened and passed through a linear layer to obtain features that are then fed to the LSTM.

Next we present a framework that combines the 3D-CNN with LSTM. The 3D-CNN acts as an encoder at sub-gesture level and provides representations for groups of few frames. These are then fed as sequences to the LSTM to get the final prediction. This can be thought of as learning a two-step hierarchy, where gestures are made up of sub-gestures. The 3D-CNN learns representations at sub-gesture level followed by the LSTM which makes prediction by building on the sub-gesture features. As many gestures have considerable overlap in their sub-gestures (raising or lowering of hand), having a two-level learning of temporal evolution of the gesture helps in better modeling the gesture. Also, all gestures will not span across the same number of frames. This is a limitation while using a CNN for classification as it requires a fixed length input. A recurrent network allows us to process arbitrary length videos.

Finally, we use our trained baseline CNN as a teacher and use its softmax output to train much smaller variants of our joint CNN and LSTM models. The benefits of this approach is demonstrated in the experiments section. In the subsequent sub-sections we detail out the architectures used in each of the aforementioned approaches.

\subsection{Baseline 3D-CNN}
\label{ssec:cnn}

\begin{figure}
  \includegraphics[width=\linewidth]{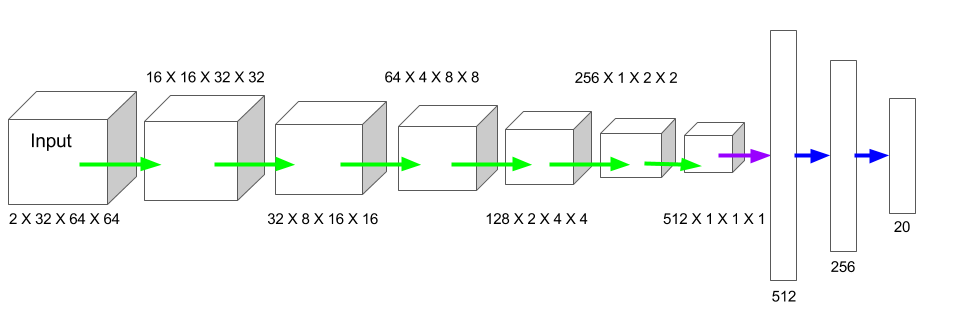}
  \caption{Baseline 3D-CNN architecture. Each block shows dimensions in the format (\textit{channels $\times$ number of frames $\times$ height $\times$ width}). Green: Conv layers, Purple: Flattening, Blue: FC layers.}
  \label{fig:3dcnn}
\end{figure}

Each gesture is represented using 32 consecutive frames, each of dimension $64 \times 64$ and two channels: depth and grayscale. For videos longer than 32 frames, the central 32 frames are extracted, while shorter videos are zero-padded. The model consists of six convolution layers, followed by two fully-connected layers (FCs). Each layer is followed by ReLU activation. Convolution layers are additionally followed by batch normalization. We do not use any pooling layer and sub-sampling is performed using padded convolution by varying the strides. For down-sampling by a factor of two, convolution is performed with filter size 4, stride 2 and padding 1 on each side. Whereas, to keep the size constant during convolution we use filter size 3, with stride 1 and padding 1 on each side. Figure \ref{fig:3dcnn} illustrates the model.

\subsection{Baseline LSTM}
\label{ssec:lstm}

Each video is divided into chunks of 4 frames of dimension $64 \times 64$, for grayscale and depth channels respectively. After flattening, it is passed through a $512$ dimensional linear layer, followed by ReLU activation. Sequences of $512$ dimensional feature vectors are fed into the LSTM, containing $256$ units, for final classification.

\subsection{Joint 3D-CNN and LSTM}
\label{ssec:joined}

\begin{figure}
  \includegraphics[width=\linewidth]{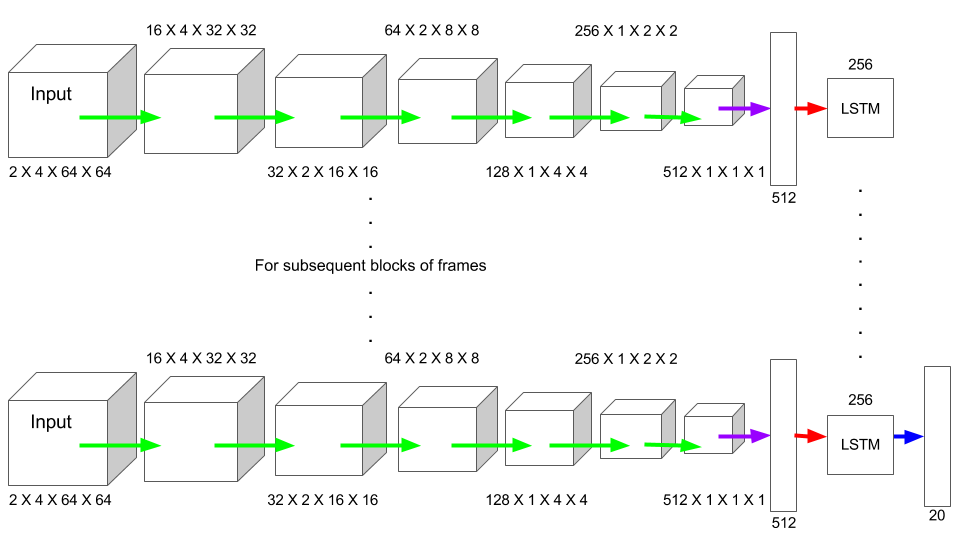}
  \caption{Joined 3D-CNN and LSTM architecture (best viewed in color). Each block shows dimensions in the format (\textit{channels $\times$ number of frames $\times$ height $\times$ width}). Green: Conv layers, Purple: Flattening, Red: LSTM connection at each time-step, Blue: FC layers.}
  \label{fig:joined}
\end{figure}

The joint model consists of an encoder 3D-CNN and an LSTM for sequence classification. The encoder takes blocks of $4 \times 64 \times 64$ frames of two channels (depth and grayscale) as input and produces a feature vector of it. The LSTM takes those vectors sequentially for an entire gesture and gives the final prediction. Architecture of the 3D-CNN is similar to the one used in our baseline model, except the number of frames in each block (4 \textit{versus} 32), and the LSTM consists of $256$ units. The complete model is shown in Figure \ref{fig:joined}, which is trained end-to-end.

\subsection{Knowledge Distillation from Baseline 3D-CNN Model to Joined Model}
\label{ssec:darkknowledge}

We train our baseline 3D-CNN model till convergence and obtain the softmax output for each training sample. As suggested in \cite{dark-knowledge}, for each gesture we obtain softened version of softmax using:
\begin{equation}\label{eq1}
  P_i = \frac{e^{\frac{Z_i}{T}}}{\sum_{j=1}^{c} e^{\frac{Z_j}{T}}}, \forall i \in \{1, ... c\},
\end{equation}
where $c$ is the number of classes and $T$ is the temperature, set depending on how ``soft'' we want the distribution to be.

We train smaller variants (referred to as \textit{medium} and \textit{small}) of our joint model obtained by reducing the number of feature maps (channels) in each layer by two and four times, respectively. The aim is to minimize the following loss function:
\begin{equation}\label{eq2}
  L = \alpha L^{(soft)} + (1-\alpha)L^{(hard)},
\end{equation}
where $L^{(soft)}$ is the cross-entropy loss between pre-trained teacher's and student's softened softmax output, $L^{(hard)}$ is the cross-entropy loss between the actual class label and model output, and $\alpha$ is a weighting parameter (set as $0.5$ in our experiments).

\section{Experiments and Results}
\label{sec:exp}

\subsection{Dataset}
\label{ssec:dataset}

\begin{figure}
  \includegraphics[width=\linewidth]{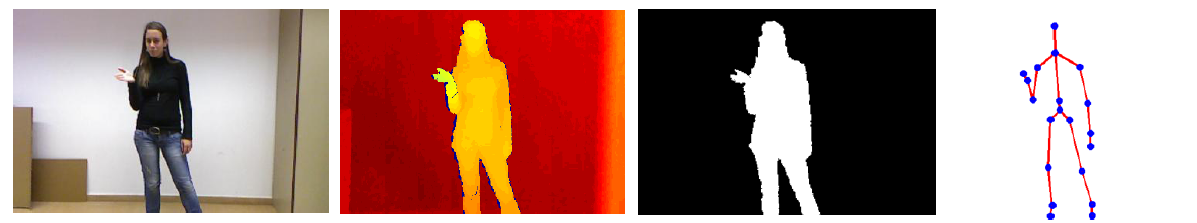}
  \caption{Example frame modalities from the dataset.}
  \label{fig:dataset}
\end{figure}

The Chalearn 2014 Looking at People Challenge (track 3) \cite{Escalera2015} dataset is a vocabulary of 20 different Italian cultural/ anthropological signs, performed by 27 different users with variations in surroundings, clothing, lighting and gesture movement. The dataset consists of 7,754 gestures for training, 3,362 gestures for validation and 2,742 gestures for testing. The videos are recorded using Microsoft Kinect and consist of RGB, depth, user mask and skeleton/joint information for each frame of video as shown in Figure \ref{fig:dataset}.

Each video may contain multiple gestures, performed one after another. The training and validation data consists of soft segmented annotation for the gestures in the videos, but no such annotation is available for the testing data. Since the main aim of this work is gesture classification, not segmentation, we use the validation data as our testing data and report accuracies on the same here-on. To maintain parity, we compare against the accuracies obtained on the validation data by other methods also. Further, we randomly choose 2,000 videos from training set to act as our validation data.

\begin{table}
  \centering 
  \begin{tabular}{| c | c |} 
  \hline
  \textbf{Input mode} & \textbf{Accuracy(\%)} \\ 
  \hline\hline
  Hand & $87.33$ \\ 
  \hline
  Upper-body & $87.59$ \\
  \hline
  Combined & $90.13$ \\
  \hline
  \end{tabular}
  \caption{Baseline 3D-CNN on different input modalities.}
  \label{table:tab1}
\end{table}

\begin{figure}
  \includegraphics[width=\linewidth]{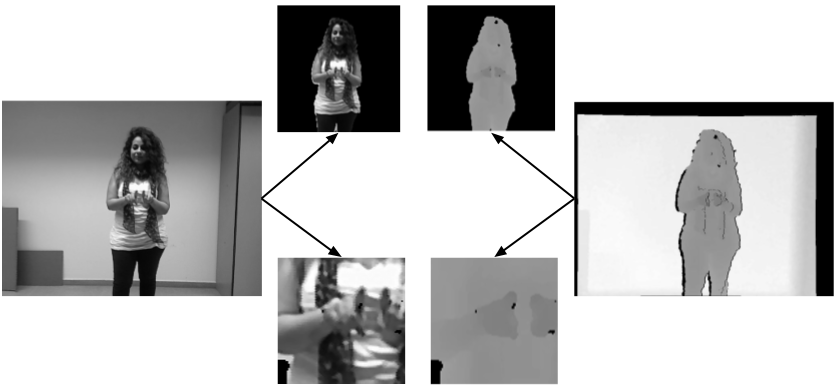}
  \caption{Left and right images are example grayscale and depth frames from the dataset. Center rows show the upper-body and hand input frames for our 	models obtained from it.}
  \label{fig:input}
\end{figure}

\begin{table}
  \centering
  \begin{tabular}{| c | c |} 
  \hline
  \textbf{Method/Model} & \textbf{Accuracy(\%)} \\ 
  \hline\hline
  Baseline LSTM & $86.6$ \\ 
  \hline
  Baseline 3D-CNN & $90.1$ \\
  \hline
  \textbf{3D-CNN + LSTM (ours)} & $\textbf{93.2}$ \\
  \hline
   Wu \textit{et al.} \cite{7423804} & $87.9$ \\ 
  \hline
  Pigou \textit{et al.} \cite{Pigou2015} & $91.4$ \\
  \hline
  Neverova \textit{et al.} \cite{DBLP:journals/corr/NeverovaWTN15} & $\textbf{96.8}$ \\
  \hline
  \end{tabular}
  \caption{Accuracies obtained using our model compared with \textit{state-of-the-art} methods.}
  \label{table:tab2}
\end{table}

\begin{table*}[t!p]
  \centering
  \begin{tabular}{| c | c | c | c | c |} 
  \hline
  & \textbf{Model} & \shortstack{\# \textbf{of parameters} \\ (in millions)} & \textbf{Trained using} & \textbf{Accuracy(\%)}\\ 
  \hline\hline
  \textit{Original} & 3D-CNN + LSTM & $18.37$ & class labels & $93.18$\\ 
  \hline
  \textit{Teacher} & 3D-CNN & $18.82$ & class labels & $90.13$\\ 
  \hline
  \multirow{4}{*}{\textit{Student}} & \multirow{2}{*}{\shortstack{3D-CNN + LSTM \\ \textit{(medium)}}} & \multirow{2}{*}{\boldmath{$4.59$}} & class labels & $86.18$ \\ \cline{4-5}
  &&& class labels and softmax output of \textit{teacher} & \boldmath{$88.35$} \\ \cline{2-5}
  & \multirow{2}{*}{\shortstack{3D-CNN + LSTM \\ \textit{(small)}}} & \multirow{2}{*}{\boldmath{$1.15$}} & class labels & $81.50$ \\ \cline{4-5}
  &&& class labels and softmax output of \textit{teacher} & \boldmath{$86.05$} \\ \cline{2-5}
  \hline
  \end{tabular}
  \caption{Knowledge Distillation from baseline 3D-CNN to CNN + LSTM.}
  \label{table:tab4}
\end{table*}

\begin{table*}
  \centering
  \begin{tabular}{| l | c | c | c | c | c |} 
  \hline
  \multirow{2}{*}{\textbf{Method}} & \multirow{2}{*}{\shortstack{\# \textbf{of parameters} \\ (in millions)}} 
  & \multicolumn{2}{|c|}{{\textbf{Single-precision}}} & \multicolumn{2}{|c|}{{\textbf{Half-precision}}}\\ \cline{3-6} 
  && \textbf{Model size} (MB) & \textbf{Accuracy(\%)} & \textbf{Model size} (MB) & \textbf{Accuracy(\%)} \\
  \hline \hline
  1.~\textit{Teacher} 3D-CNN & $18.82$ & $72$ & $90.13$ & $36$ & $89.5$ \\ \hline
  2.~\textit{Original} 3D-CNN + LSTM & $18.37$ & $71$ & $93.18$ & $35.5$ & $93.18$ \\ \hline
  3.~\textit{Student} 3D-CNN + LSTM & $1.15$ & $4.5$ & $86.05$ & $2.25$ & $85.98$ \\ \hline
  4.~\textbf{Sparse model of (3)} & \boldmath{$0.25$} & \boldmath{$1.12$} & \boldmath{$86.05$} & \boldmath{$0.635$} & \boldmath{$85.98$} \\ \hline
  \end{tabular}
  \caption{Reduction is size along with performance impact of the student model and sparse model.}
  \label{table:tab5}
\end{table*}

\subsection{Input}
\label{ssec:input}

For each video frame we use the depth and convert the RGB frames to grayscale and concatenate them to obtain two-channel inputs for our models. Since all the gestures occur in the upper-body region, we extract it using the skeleton information and resize it to $64 \times 64$. Thus inputs to our models are of dimension $2 \times T \times 64 \times 64$ , where T is $32$ for the baseline 3D-CNN model and $4$ for the joint model (as mentioned in section \ref{ssec:cnn} and \ref{ssec:joined} respectively).

Similar to \cite{Pigou2015}, using skeleton information, we crop out the highest hand region for each gesture, resize to $64 \times 64$ and use them as input. Table \ref{table:tab1} shows accuracies obtained with our baseline 3D-CNN model while using hand, upper-body and both combined, as input. Since the combination of upper-body and hand gives the best performance, all following experiments use that as input (Figure \ref{fig:input}). We also perform rotation, translation and zooming on the frames for data augmentation.

\subsection{Results and Analysis}
\label{ssec:result}

Table \ref{table:tab2} shows the accuracies obtained using our models. The baseline LSTM performs the worst since the spatial and temporal information are not conserved while flattening out. Baseline 3D-CNN preserves such information across dimensions and performs better, but it is not suitable for capturing long-term dependencies. The best accuracy is obtained using the joint model which takes advantage of both 3D-CNN, for feature learning, and LSTM, for classification.

We compare our CNN + LSTM model against other \textit{state-of-the-art} methods on the same dataset in Table \ref{table:tab2}. Wu \textit{et al.} \cite{7423804} described a novel method called Deep Dynamic Neural Networks, a semi-supervised hierarchical dynamic framework based on a Hidden Markov Model (HMM) and learned high-level spatio-temporal representations using a Deep Belief Network (DBN) and 3D-CNN, suited to the input modality. Pigou \textit{et al.} \cite{Pigou2015} used an approach similar to our baseline 3D-CNN approach, but their architecture is very different from our CNN. ModDrop \cite{DBLP:journals/corr/NeverovaWTN15} used a multi-scale and multi-modal deep learning approach. It performed careful initialization of individual modalities and fused multiple modalities at several spatial and temporal scales. Their architecture is a big cascade having individual branches for each hand, each modalities, and articulated pose based features. Our simpler architecture works directly with the raw pixels, without the need to perform any pre-training or initializations.

The effect of Knowledge Distillation is analyzed in Table \ref{table:tab4}. Reducing the number of parameters limits the model's capacity to learn from the class labels directly, but it can be alleviated when trained according to Equation \ref{eq2}, using soft outputs from a larger pre-trained teacher network. We get almost at-par accuracy with the teacher, but at a much reduced training cost, making it both computationally and memory efficient. Further, training with Adam optimizer makes most of the parameters of the student have very low weights, which allows for further compression to obtain a sparse model. Removing weights having magnitude below $2^{-100}$ got rid of $\sim905K$ parameters out of $1.15 M$, of our small student network with no drop in performance. Table \ref{table:tab5} compares this sparse model with our other models.

\section{Conclusion}
\label{sec:conclusion}

We show how joint 3D-CNN and LSTM model for gesture recognition from videos, leverages the best of both 3D convolution and recurrent network to model the sequential evolution of information in a video, while allowing to process arbitrary length videos. 

We also show how information can be distilled from a larger model to models with $16\times$ and $4\times$ fewer parameters. Size of models could be further reduced using a sparse representation as discussed above. This not only benefits training time but also makes it possible to use them in low-memory and low-power embedded devices.

\newpage
\label{sec:ref}
\begin{small}
  \bibliographystyle{IEEEbib}
  \bibliography{paper,refs}
\end{small}

\end{document}